\documentclass[11pt,a4paper]{article}
\usepackage[hyperref]{emnlp2018}
\usepackage{times}
\usepackage{latexsym}
\usepackage[utf8]{inputenc}
\usepackage{times}
\usepackage{latexsym}
\usepackage{booktabs}
\usepackage{amsmath}
\usepackage{float}
\usepackage{amssymb}
\usepackage{graphicx}
\usepackage{tabularx}

\newcolumntype{b}{>{\hsize=1.3\hsize}X}
\newcolumntype{s}{>{\hsize=.7\hsize}X}

\usepackage{url}
\usepackage{multirow}
\usepackage[section]{placeins}

\aclfinalcopy 

\begin{document}
\title{Predictive Embeddings for Hate Speech Detection on Twitter}

\author{Rohan Kshirsagar$^1$  Tyrus Cukuvac$^1$  Kathleen McKeown$^1$  Susan McGregor$^2$\\
		$^1$Department of Computer Science at Columbia University \\
        $^2$School of Journalism at Columbia University \\
	    {\tt rmk2161@columbia.edu thc2125@columbia.edu} \\
        {\tt kathy@cs.columbia.edu sem2196@columbia.edu}  
  }

\maketitle
\begin{abstract}
We present a neural-network based approach to classifying online hate speech in general, as well as racist and sexist speech in particular. Using pre-trained word embeddings and max/mean pooling from simple, fully-connected transformations of these embeddings, we are able to predict the occurrence of hate speech on three 
commonly used
publicly available datasets. Our models match or outperform state of the art F1 performance on all three datasets 
using significantly fewer parameters and minimal feature preprocessing compared to previous methods.

\end{abstract}
\section{Introduction}
The increasing popularity of social media platforms like Twitter for both personal and political communication \cite{knight-public-square} has  seen a well-acknowledged rise in the presence of toxic and abusive speech on these platforms \cite{national-review-cesspool,mother-jones-cesspool}. 
Although the terms of services on these platforms typically forbid hateful and harassing speech, enforcing these rules has proved challenging, as identifying hate speech speech at scale is still a largely unsolved problem in the NLP community. 
\citet{waseem2016hateful}, for example, identify many ambiguities in classifying abusive communications, and highlight the difficulty of clearly defining the parameters of such speech. This problem is compounded by the fact that identifying abusive or harassing speech is a challenge for humans as well as automated systems.\par
%
%
Despite the lack of consensus around what constitutes abusive speech, \textit{some} definition of hate speech must be used to build automated systems to address it. 
We rely on \citet{davidson2017automated}'s definition of hate speech,
 specifically:
``language that is used to express hatred towards a targeted group or is intended to be derogatory, to humiliate,
or to insult the members of the group.''\par 
In this paper, we present a neural classification system that uses minimal preprocessing to take advantage of a modified Simple Word Embeddings-based Model~\citep{shen2018on} to predict the occurrence of hate speech. Our classifier features:
\begin{itemize}
\item A simple deep learning approach with few parameters enabling quick and robust training
\item Significantly better performance than two other state of the art methods on publicly available datasets
\item An interpretable approach facilitating analysis of results
\end{itemize}
\par
In the following sections, we discuss related work on hate speech classification, followed by a description of the datasets, methods and results of our study.
%
%
%
%
%
%
\section{Related Work}
Many efforts have been made to classify hate speech using data scraped from online message forums and popular social media sites such as Twitter and Facebook. 
%
%
%
%
\citet{waseem2016hateful} applied a logistic regression model that used one- to four-character n-grams for classification of tweets labeled as racist, sexist or neither.
\citet{davidson2017automated} experimented in classification of hateful as well as offensive but not hateful tweets. They applied a logistic regression classifier with L2 regularization using word level n-grams and various part-of-speech, sentiment, and tweet-level metadata features. \par 
Additional projects have built upon the data sets created by Waseem and/or Davidson. For example, \citet{park17} used a neural network approach with two binary classifiers: one to predict the presence abusive speech more generally, and another to discern the form of abusive speech.\par
\citet{zhang2018}, meanwhile, used pre-trained word2vec embeddings, which were then fed into a convolutional neural network (CNN) with max pooling to produce input vectors for a Gated Recurrent Unit (GRU) neural network. 
Other researchers have experimented with using metadata features from tweets. \citet{founta2018unified} built a classifier composed of two separate neural networks, one for the text and the other for metadata of the Twitter user, that were trained jointly in interleaved fashion. 
Both networks used in combination - and especially when trained using transfer learning -
achieved higher F1 scores than either neural network classifier alone.\par
In contrast to the methods described above, our approach relies on a simple word embedding (SWEM)-based architecture \citep{shen2018on}, reducing the number of required parameters and length of training required, while still yielding improved performance and resilience across related classification tasks. 
Moreover, our network is able to learn flexible vector representations that demonstrate associations among words typically used in hateful communication. Finally, while metadata-based augmentation is intriguing, here we sought to develop an approach that would function well even in cases where such additional data was missing due to the deletion, suspension, or deactivation of accounts. 
\par
%
%
%
%
\section{Data}
In this paper, we use three data sets
from the literature
to train and evaluate our own classifier. Although all address the category of hateful speech, they used different strategies of labeling the collected data.
%
%
Table~\ref{Tab:datasets} shows the characteristics of the datasets.\par
Data collected by \citet{waseem2016hateful}, which we term the {\bf Sexist/Racist (SR)} data set\footnote{Some Tweet IDs/users have been deleted since the creation, so the total number may differ from the original}, was collected using
an initial Twitter search followed by analysis and filtering by the authors and their team who identified 
17 common phrases, hashtags, and users that were indicative of abusive speech. 
\citet{davidson2017automated} collected the {\bf HATE dataset} by searching for tweets using a lexicon provided by \textit{Hatebase.org}. 
The final data set we used, which we call {\bf HAR}, was collected by \citet{golbeck2017large};  we removed all retweets reducing the dataset to 20,000 tweets. 
Tweets were labeled as ``Harrassing'' or ``Non-Harrassing'';
hate speech was not explicitly labeled, but treated as an unlabeled subset of the broader ``Harrassing'' category\citep{golbeck2017large}. \par
\begin{table}
\tabcolsep=0.1cm
\begin{center}
\begin{tabular}{lrrrr}
\toprule
Dataset  & \multicolumn{3}{l}{Labels and Counts} & Total\\
\midrule
\multirow{2}{4em}{SR}   & Sexist & Racist                 & Neither                            \\
                        &  3086  & 1924                   & 10,898           & \textbf{15,908} \\
\midrule
\multirow{2}{4em}{HATE} & \multicolumn{2}{r}{Hate Speech} & Not Hate Speech\\
                        & \multicolumn{2}{r}{1430}        & 23,353           & \textbf{24,783} \\
\midrule
\multirow{2}{4em}{HAR}  & \multicolumn{2}{r}{Harassment}  & Non Harassing\\
                        & \multicolumn{2}{r}{5,285}       & 15,075           & \textbf{20,360} \\
\bottomrule
\end{tabular}
\end{center}
\caption{Dataset Characteristics}
\label{Tab:datasets}
\end{table}
\section {Transformed Word Embedding Model (TWEM)}
Our training set consists of $N$ examples $\{X^i, Y^i\}_{i=1}^N$ where the input $X^i$ is a sequence of tokens $w_1, w_2, ..., w_T$, and the output $Y^i$ is 
the numerical class for the hate speech class. 
Each input instance represents a Twitter post and thus, is not limited to a single sentence. \par 

We modify the SWEM-concat \citep{shen2018on} architecture to allow better handling of infrequent and unknown words and to capture non-linear word combinations. \par

\subsection{Word Embeddings}
 Each token in the input is mapped to an embedding. We used the 300 dimensional embeddings for all our experiments, so each word $w_t$ is
 mapped to $x_t \in {\mathbb R}^{300}$. We denote the full embedded sequence as $x_{1:T}$. 
 We then transform each word embedding by applying 
 300 dimensional 1-layer Multi Layer Perceptron (MLP) $W_t$ with a Rectified Liner Unit (ReLU) activation to form an updated embedding space $z_{1:T}$. We find this better handles unseen or rare tokens in our training data by projecting the pretrained embedding into a space that the encoder can understand.
 
\subsection{Pooling}
We make use of two pooling methods on the updated embedding space $z_{1:T}$. We employ a max pooling operation on $z_{1:T}$ to capture salient word features from our input; this representation is denoted as $m$. This forces words that are highly indicative of hate speech to higher positive values within the updated embedding space. We also average the embeddings $z_{1:T}$ to capture the overall meaning of the sentence, denoted as $a$, which provides a strong conditional factor in conjunction with the max pooling output. This also helps regularize gradient updates from the max pooling operation. 

\subsection{Output}

We concatenate $a$ and $m$ to form a document representation 
$d$ and feed the representation into a 50 node 2 layer MLP followed by  ReLU Activation to allow for increased nonlinear representation learning. This representation forms the preterminal layer and is 
passed to a fully connected softmax layer whose output is the probability distribution over labels. 

\section{Experimental Setup}
We tokenize the data using Spacy \citep{honnibal-johnson:2015:EMNLP}.
We use 300 Dimensional Glove Common Crawl Embeddings (840B Token) \citep{pennington2014} and fine tune them for the task. 
We experimented extensively with  pre-processing variants and our results showed better performance without lemmatization and lower-casing (see supplement for details).
We pad each input to 50 words. We train using RMSprop with a learning rate of .001 and a batch size of 512. We add dropout with a drop rate of 0.1 in the final layer 
to reduce overfitting \citep{srivastava2014dropout}, batch size, and input length empirically through random hyperparameter search.

\par All of our results are produced from 10-fold cross validation to allow comparison with previous results.
We trained a logistic regression baseline model (line 1 in
Table~\ref{table:results}) using character ngrams  and word unigrams using TF*IDF weighting~\citep{salton1987term}, to provide a baseline since HAR has no reported results. For the SR and HATE datasets, the authors reported their trained best logistic regression model's\footnote{Features described in Related Works section} results on their respective datasets.

\begin{table}[t]
\tabcolsep=0.05cm
\begin{tabular}{lrrrrr}
\toprule
\textbf{Method}  & \textbf{SR} & \textbf{HATE} & \textbf{HAR} \\
\midrule
LR(Char-gram + Unigram)  & 0.79  & 0.85  &0.68   \\
\midrule
LR\citep{waseem2016hateful} & 0.74 & -  & -\\
\midrule
LR \citep{davidson2017automated}  & -  & 0.90  & -  \\
\midrule
CNN \citep{park17} & 0.83 & - & - \\
\midrule
GRU Text \citep{founta2018unified}  & 0.83 & 0.89 & - \\

GRU Text + Metadata &  \textbf{0.87} & 0.89 & - \\
\midrule
TWEM (Ours) & 0.86 & \textbf{0.924} & \textbf{0.71 }\\

\bottomrule

\end{tabular}
\caption{F1 Results\footnotemark}
\label{table:results}
\end{table}
\footnotetext{SR: Sexist/Racist \citep{waseem2016hateful}, HATE: Hate \citep{davidson2017automated} HAR: Harassment \citep{golbeck2017large}}
\section{Results and Discussion}
The approach we have developed establishes a new state of the art
for classifying hate speech, outperforming previous results by as much as 12 F1 points.  Table \ref{table:results} illustrates the robustness of our method, 
which often outperform previous results, measured by weighted F1. \footnote{This was used in previous work, as confirmed by checking with authors}

Using the Approximate Randomization (AR) Test \citep{riezler2005some}, we perform significance testing using a 75/25 train and test split

to compare against \citep{waseem2016hateful} and \citep{davidson2017automated}, whose models we re-implemented. We found 0.001 significance compared to both methods.
We also include in-depth precision and recall results for all three datasets in the supplement. 


Our results indicate better performance than several more complex approaches, including
\citet{davidson2017automated}'s best model (which used word and part-of-speech ngrams, sentiment, readability, text, and Twitter specific features), \citet{park17} (which used two fold classification and a hybrid of word and character CNNs, using approximately twice the parameters we use excluding the word embeddings) and even recent work by \citet{founta2018unified}, (whose best model relies on GRUs, metadata including popularity, network reciprocity, and subscribed lists). 

On the SR dataset, we outperform \citet{founta2018unified}'s text based model by 3 F1 points, while just falling short of the Text + Metadata Interleaved Training model. While we appreciate the potential added value of metadata, we believe a tweet-only classifier has merits because retrieving features from the social graph is not always tractable in production settings. Excluding the embedding weights,  our model requires ~100k parameters , while \citet{founta2018unified} requires 250k parameters.

\subsection{Error Analysis}
\textbf{False negatives\footnote{Focused on the SR Dataset \citep{waseem2016hateful}}} \newline
Many of the false negatives we see are
specific references to characters in the TV show ``My Kitchen Rules'', rather than something about women in general. Such examples may be innocuous in isolation but could potentially be sexist or racist in context. While this may be a limitation of considering only the content of the tweet, it could also be a mislabel. 
\begin{quote}
Debra are now my most hated team on \#mkr after least night's ep. Snakes in the grass those two.
 \end{quote}
Along these lines, we also see correct predictions of innocuous 
speech, but find data mislabeled as hate speech: 
\begin{quote}
@LoveAndLonging ...how is that example "sexism"?
\end{quote}
\begin{quote}
@amberhasalamb ...in what way? 
\end{quote}
Another case our classifier misses is problematic speech within a hashtag: 
\begin{quote}
:D @nkrause11 Dudes who go to culinary school: \#why \#findawife \#notsexist :) 
\end{quote}
This limitation could be potentially improved through the use of character convolutions or subword tokenization.\\
\textbf{False Positives}\newline
In certain cases, our model seems to be learning user names instead of semantic content:
\begin{quote}
RT @GrantLeeStone: @MT8\_9 I don't even know what that is, or where it's from. Was that supposed to be funny? It wasn't. 
\end{quote}

Since the bulk of our model's weights are in the embedding and embedding-transformation matrices, we cluster the SR vocabulary using these transformed embeddings to clarify our intuitions about the model (\ref{table:clusters}). We elaborate on our clustering approach in the supplement.  We see that the model learned general semantic groupings of words associated with hate speech as well as specific idiosyncrasies related to the dataset itself (e.g. \textit{katieandnikki})
\begin{table}[h]
\scalebox{0.72}{
\begin{tabular}{p{3cm}|p{7cm}}
\hline
\textbf{Cluster}                                                                                                                             & \textbf{Tokens}                                                                                                                                                                                                                             \\ \hline
Geopolitical and religious references around Islam and the Middle East                                                                         & bomb, mobs, jewish, kidnapped, airstrikes, secularization, ghettoes, islamic, burnt, murderous, penal, traitor, intelligence, molesting, cannibalism                                                  \\ \hline
Strong epithets and adjectives associated with harassment and hatespeech                                                                     & liberals, argumentative, dehumanize, gendered, stereotype, sociopath,bigot, repressed, judgmental, heinous, misandry, shameless, depravity, scumbag,                         \\ \hline
Miscellaneous                                  & turnt, pedophelia, fricken, exfoliated, sociolinguistic, proph, cissexism, guna, lyked, mobbing, capsicums, orajel, bitchslapped, venturebeat, hairflip, mongodb, intersectional, agender                                              \\ \hline
Sexist related epithets and hashtags& malnourished, katieandnikki,   chevapi, dumbslut, mansplainers, crazybitch, horrendousness, justhonest, bile, womenaretoohardtoanimate,  \\ \hline
Sexist, sexual, and pornographic terms                                                                                                       & actress, feminism, skank, breasts, redhead, anime, bra, twat, chick, sluts, trollop, teenage, pantyhose, pussies, dyke, blonds,                                                                                 \\ \hline
\end{tabular}}
\caption{Projected Embedding Cluster Analysis from SR Dataset}
\label{table:clusters}
\end{table}
\section{Conclusion}
Despite minimal tuning of hyper-parameters, fewer weight parameters, minimal text preprocessing, and no additional metadata, the model performs remarkably well on standard hate speech datasets. Our clustering analysis adds interpretability enabling inspection of results.

Our results indicate that the majority of recent deep learning models in hate speech may rely on word embeddings for the bulk of predictive power and the addition of sequence-based parameters provide minimal utility. Sequence based approaches are typically important when phenomena such as negation, co-reference, and context-dependent phrases are salient in the text and thus, we suspect these cases are in the minority for publicly available datasets. 
We think it would be valuable to study the occurrence of such linguistic phenomena in existing datasets and construct new datasets that have a better representation of subtle forms of hate speech.
In the future, we plan to investigate character based representations, using character CNNs and highway layers \citep{kim2016character} along with word embeddings to allow robust representations for sparse words such as hashtags. 

\newpage
\bibliographystyle{acl_natbib_nourl}
\bibliography{hatespeech_detection}
\newpage
\appendix
\section{Supplemental Material}

We experimented with several different preprocessing variants and were surprised to find that reducing preprocessing improved the performance on the task for all of our tasks. We go through each preprocessing variant with an example and then describe our analysis to compare and evaluate each of them. 
\subsection {Preprocessing}
\textbf{Original text} \newline
\begin{quote}
RT @AGuyNamed\_Nick Now, I'm not sexist in any way shape or form but I think women are better at gift wrapping. It's the XX chromosome thing
\end{quote}
\textbf{Tokenize (Basic Tokenize: Keeps case and words intact with limited sanitizing)}
\begin{quote}
RT @AGuyNamed\_Nick Now , I 'm not sexist in any way shape or form but I think women are better at gift wrapping . It 's the XX chromosome thing
\end{quote}
\textbf{Tokenize Lowercase: Lowercase the basic tokenize scheme}
\begin{quote}
rt @aguynamed\_nick now , i 'm not sexist in any way shape or form but i think women are better at gift wrapping . it 's the xx chromosome thing
\end{quote}
\textbf{Token Replace: Replaces entities and user names with placeholder)}
\begin{quote}
ENT USER now , I 'm not sexist in any way shape or form but I think women are better at gift wrapping . It 's the xx chromosome thing
\end{quote}
\textbf{Token Replace Lowercase: Lowercase the Token Replace Scheme}
\begin{quote}
ENT USER now , i 'm not sexist in any way shape or form but i think women are better at gift wrapping . it 's the xx chromosome thing
\end{quote}

We did analysis on a validation set across multiple datasets to find that the "Tokenize" scheme was by far the best. We believe that keeping the case in tact provides useful information about the user. For example, saying something in all CAPS is a useful signal that the model can take advantage of. 
\begin{table}
\centering
\label{my-label}
\begin{tabular}{lr}
\textbf{Preprocessing Scheme}   & \textbf{Avg. Validation Loss} \\
Token Replace Lowercase          & 0.47                   \\
Token Replace & 0.46                  \\
Tokenize                 & 0.32                  \\
Tokenize Lowercase          & 0.40                 
\end{tabular}
\caption{Average Validation Loss for each Preprocessing Scheme}
\end{table}

\subsection{In-Depth Results}
\begin{table}[h]
\vspace{.7cm}
\tabcolsep=0.17cm
\begin{tabular}{lrrrrrr}
		\toprule
       & \multicolumn{3}{c}{Waseem \citeyear{waseem2016hateful}} & \multicolumn{3}{c}{Ours}  \\
       \midrule
       & P       & R       & F1        & P & R & F1   \\
       \midrule
none   & 0.76            & 0.98         & 0.86      & 0.88      & 0.93   & \textbf{0.90}  \\
sexism & 0.95            & 0.38         & 0.54      & 0.79      & 0.74   & \textbf{0.76} \\
racism & 0.85            & 0.30          & 0.44     & 0.86      & 0.72   & \textbf{0.78} \\
&&&0.74&&&\textbf{0.86} \\
\bottomrule
\end{tabular}
\caption{SR Results}
\label{table:waseem}

\tabcolsep=0.13cm
\vspace{.7cm}
\begin{tabular}{lrrrrrr}
\toprule
       & \multicolumn{3}{c}{Davidson \citeyear{davidson2017automated}} & \multicolumn{3}{c}{Ours}  \\
       \midrule
       & P       & R       & F1        & P & R & F1   \\
       \midrule
none   & 0.82            & 0.95         & 0.88      & 0.89      & 0.94   & \textbf{0.91}  \\
offensive & 0.96            & 0.91         & 0.93      & 0.95      & 0.96   & \textbf{0.96} \\
hate & 0.44            & 0.61          & \textbf{0.51}     & 0.61      & 0.41   &0.49 \\
&&&0.90&&&\textbf{0.924} \\
\bottomrule
\end{tabular}
\caption{HATE Results}
\label{table:davidson}
\vspace{.5cm}
\tabcolsep=0.22cm
\begin{tabular}{lrrrr}
\toprule
Method                       & Prec & Rec & F1    & Avg F1 \\ 
\midrule
Ours             & 0.713     & 0.206  &\textbf{ 0.319} & \textbf{0.711}\\
LR Baseline & 0.820     & 0.095  & 0.170 & 0.669   \\
\bottomrule

\label{table:golbeck}
\end{tabular}
\caption{HAR Results}
\end{table}
\subsection{Embedding Analysis}
Since our method was a simple word embedding based model, we explored the learned embedding space to analyze results. For this analysis, we only use the max pooling part of our architecture to help analyze the learned embedding space because it encourages salient words to increase their values to be selected. We projected the original pre-trained embeddings to the learned space using the time distributed MLP. We summed the embedding dimensions for each word 
and sorted by the sum in descending order to find the 1000 most salient word embeddings from our vocabulary. We then ran PCA \citep{jolliffe1986principal} to reduce the dimensionality of the projected embeddings from 300 dimensions to 75 dimensions. This captured about 60\% of the variance. Finally, we ran K means clustering for $k=5$ clusters to organize the most salient embeddings in the projected space. 

The learned clusters from the SR vocabulary were very illuminating (see Table \ref{table:clusters}); they gave insights to how hate speech surfaced in the datasets. One clear grouping we found is the misogynistic and pornographic group, which contained words like {\em breasts}, {\em blonds}, and {\em skank}. 
Two other clusters had references to geopolitical and religious issues in the Middle East and disparaging and resentful epithets that could be seen as having an intellectual tone. This hints towards the subtle pedagogic forms of hate speech that surface.  

We ran silhouette analysis \citep{scikit-learn} on the learned clusters to find that the clusters from the learned representations
had a 35\% higher silhouette coefficient using the projected embeddings compared to 
the clusters created from 
the original pre-trained embeddings. This reinforces the claim that our training process pushed  hate-speech related words together, and words from other clusters further away, thus, structuring the embedding space effectively for detecting hate speech. 

\begin{table}[h]
\scalebox{0.72}{
\begin{tabular}{p{3cm}|p{7cm}}
\hline
\textbf{Cluster}                                                                                                                             & \textbf{Tokens}                                                                                                                                                                                                                             \\ \hline
Geopolitical and religious references around Islam and the Middle East                                                                         & bomb, mobs, jewish, kidnapped, airstrikes, secularization, ghettoes, islamic, burnt, murderous, penal, traitor, intelligence, molesting, cannibalism                                                  \\ \hline
Strong epithets and adjectives associated with harassment and hatespeech                                                                     & liberals, argumentative, dehumanize, gendered, stereotype, sociopath,bigot, repressed, judgmental, heinous, misandry, shameless, depravity, scumbag,                         \\ \hline
Miscellaneous                                  & turnt, pedophelia, fricken, exfoliated, sociolinguistic, proph, cissexism, guna, lyked, mobbing, capsicums, orajel, bitchslapped, venturebeat, hairflip, mongodb, intersectional, agender                                              \\ \hline
Sexist related epithets and hashtags& malnourished, katieandnikki,   chevapi, dumbslut, mansplainers, crazybitch, horrendousness, justhonest, bile, womenaretoohardtoanimate,  \\ \hline
Sexist, sexual, and pornographic terms                                                                                                       & actress, feminism, skank, breasts, redhead, anime, bra, twat, chick, sluts, trollop, teenage, pantyhose, pussies, dyke, blonds,                                                                                 \\ \hline
\end{tabular}}
\caption{Projected Embedding Cluster Analysis from SR Dataset}
\label{table:clusters}
\end{table}
\end{document}